%% file: main.tex
\RequirePackage[T1]{fontenc}
\documentclass[journal]{IEEEtran}
\usepackage{amsmath,amssymb,booktabs,cite,xurl}
\usepackage[hidelinks]{hyperref}
\hypersetup{pdftitle={When Sensors Fail: Coverage Auditing with Incomplete Sensor Records},pdfauthor={Chung-I Huang and Wen-Yi Chang}}
\begin{document}
\title{When Sensors Fail: Coverage Auditing with Incomplete Sensor Records}
\author{Chung-I Huang and Wen-Yi Chang%
\thanks{Chung-I Huang is with the Department of Management Information Systems, National Chung Hsing University, Taichung 402, Taiwan (e-mail: schumi@nchu.edu.tw).}%
\thanks{Wen-Yi Chang is with the National Center for High-Performance Computing (NCHC), National Institutes of Applied Research (NIAR), Hsinchu 300092, Taiwan (e-mail: 0303106@niar.org.tw). \textit{(Corresponding author: \mbox{Wen-Yi Chang}.)}}}
\maketitle
\begin{abstract}
\input{abstract_text}
\end{abstract}

\section{Introduction}
Prediction intervals are often evaluated by pooling forecasts across normal operation and sensor outages. That aggregation can conceal a coverage deficit associated with an outage state. A further difficulty arises when the future reading itself is unavailable: an issued forecast then has no original observation against which its interval can be scored. A reliability statement must therefore specify both the information available at forecast issue time and the availability of scoring truth.

We develop a measurement-record audit connecting these two availability conditions to interval coverage and its supporting evidence. The construction retains a complete forecast ledger, groups forecasts by issue-time outage duration, and accumulates covered and scored counts on the measurement clock. Simultaneous count bounds yield a time-uniform interval for a cumulative conditional expected coverage ratio. This is an application of established confidence-sequence theory to incomplete, dependent sensor records, rather than a new general confidence-sequence theorem.

Conformal prediction supplies marginal coverage under exchangeability \cite{vovk2005algorithmic}. Such a guarantee does not automatically provide coverage conditional on an outage state; distributional mismatch and dependence require additional care \cite{barber2023beyond}. Confidence sequences instead provide time-uniform inference for suitably adapted processes \cite{howard2021time}. Our contribution is the measurement-time accounting that connects this inference to repeated forecasts, unavailable truth, and explicit outage-specific evidence states.

\section{Outage State and Scoring Population}
\subsection{Issue-time state and original truth}
For sensor $i$ at forecast origin $t$, let $\tau_{t,i}$ be the consecutive number of unavailable readings through $t$, computed using only the current and preceding availability record. The states are observed ($\tau=0$), short ($1\leq\tau\leq3$), medium ($4\leq\tau\leq12$), and long ($\tau>12$). Runs are not restarted at input-window or calendar-day boundaries. Initial runs whose earlier history is unknown are left-censored. A duration measured in sampling steps has different clock-time meanings in different networks.

A forecast key $j$ identifies a sensor, an origin, and a horizon. Let $C_j$ be its issued prediction interval, $Y_j$ its target, and $O_j$ the indicator that the original released target is observed. Missing targets are not replaced with imputations for scoring. For an issue-time state $A$, define
\begin{equation}
\begin{aligned}
c_A^{\rm obs}&=P(Y\in C\mid A,O=1),\\
\widehat c_A&=\frac{\sum_{j\in A}O_j\mathbf 1\{Y_j\in C_j\}}
{\sum_{j\in A}O_j}.
\end{aligned}
\label{eq:observed}
\end{equation}
The empirical ratio is undefined without scored support. With several fixed fitted models, we average their covered indicators at each shared key before pooling; fitting seeds are not additional independent measurement events.

Marginal scored coverage is a mixture over states. Consequently, an infrequent or poorly observed outage state can have low coverage even when the marginal value is near the nominal level. We retain issued and scored counts separately, alongside interval widths in the original measurement units. Native sensor runs, simultaneous network-dark periods, and calendar blocks describe distinct aspects of the available evidence.

\subsection{What missing truth leaves unidentified}
Let $v_A=P(O=1\mid A)$ and let $c_A^{\rm unobs}$ denote coverage on unavailable targets. Then
\begin{equation}
\begin{aligned}
c_A^{\rm all}&=v_Ac_A^{\rm obs}+(1-v_A)c_A^{\rm unobs},\\
c_A^{\rm all}&\in[v_Ac_A^{\rm obs},\ v_Ac_A^{\rm obs}+1-v_A].
\end{aligned}
\label{eq:identification}
\end{equation}
This is an identification range under unrestricted unobserved-target coverage. It is not a confidence interval. A high observed-target coverage value cannot close this range when most issued targets are unavailable.

\section{A Time-Uniform Measurement-Record Audit}
\subsection{Whole-network measurement batches}
At measurement tick $t$, collect all previously issued forecasts whose target is released at that tick. Let $M_{A,t}$ count the originally observed targets of forecast keys in state $A$, and let $H_{A,t}$ be the corresponding covered count, averaged over the fixed fitted models. Different forecasts of the same target remain distinct keys. With $d$ sensors and $h$ forecast horizons,
\begin{equation}
0\leq H_{A,t}\leq M_{A,t}\leq B=hd.
\end{equation}
The experiments below use $h=12$. Every measurement tick is retained, including zero-support ticks and terminal horizons. Shared targets and concurrent sensors are not treated as independent observations.

Let $\mathcal F_t$ contain the frozen training and calibration state and the released measurement history through tick $t$. Intervals and state assignments must precede their target measurements. Future target availability must not select earlier forecasts or online updates. Define
\begin{equation}
C_{A,T}=\frac{\sum_{t\leq T}E(H_{A,t}\mid\mathcal F_{t-1})}
{\sum_{t\leq T}E(M_{A,t}\mid\mathcal F_{t-1})},
\label{eq:target}
\end{equation}
when the denominator is positive. This cumulative conditional expected count ratio allows coverage and observability to change with history. It differs from both the empirical ratio and the population quantity $c_A^{\rm obs}$ in \eqref{eq:observed}.

\subsection{Simultaneous count bounds and ratio projection}
For $Q\in\{H,M\}$, set $X_t^Q=Q_{A,t}/B$, choose a predictable center $p_t^Q\in[0,1]$, and write
\begin{equation}
Z_T^Q=\sum_{t\leq T}X_t^Q,\qquad
V_T^Q=\sum_{t\leq T}(X_t^Q-p_t^Q)^2.
\end{equation}
One choice is $p_1^Q=0$ and $p_t^Q=X_{t-1}^Q$. For fixed $\lambda_j\in(0,1)$ and positive weights summing to one, a bounded empirical-Bernstein construction \cite{howard2021time} gives
\begin{equation}
b(v)=\min_j\frac{\log\{4K/(\gamma w_j)\}+\psi(\lambda_j)v}{\lambda_j},
\label{eq:boundary}
\end{equation}
where $\psi(\lambda)=-\log(1-\lambda)-\lambda$, $K$ is the number of prespecified outage states, and $\gamma$ is the audit family error probability. The grid and its weights are fixed before the audit. The minimum combines error-allocated boundaries; it is not unrestricted data-driven selection of a favorable tuning parameter.

For each normalized count, form
\begin{equation}
\ell_T^Q=\max\{0,Z_T^Q-b(V_T^Q)\},\qquad
u_T^Q=\min\{T,Z_T^Q+b(V_T^Q)\}.
\end{equation}
Project the joint count rectangle to
\begin{equation}
L_{A,T}=\max\{0,\ell_T^H/u_T^M\},\qquad
U_{A,T}=\min\{1,u_T^H/\ell_T^M\}.
\end{equation}
Take $U_{A,T}=1$ when $\ell_T^M\leq0$ and return $[0,1]$ for an undefined or infeasible projection.

To see the guarantee, apply the bounded exponential-supermartingale construction separately to each count tail. Ville's inequality bounds crossing for a fixed grid component; allocating $\gamma w_j/(4K)$ to that component and summing over two counts, two tails, states, and grid components produces simultaneous count coverage. On this event the true expected numerator and denominator lie in their count rectangle. Their ratio therefore lies in its projection whenever expected scored support is positive. Thus
\begin{equation}
P\{\forall A,T:\ L_{A,T}\leq C_{A,T}\leq U_{A,T}\}\geq1-\gamma,
\label{eq:guarantee}
\end{equation}
where the quantification concerns times with positive expected support. Dependence across sensors, horizons, states, and ticks is permitted by this bounded adapted-process construction. The family is fixed for a particular dataset, prediction procedure, and nominal interval level; the guarantee does not jointly cover every experimental comparison.

\subsection{Evidence states and operational comparisons}
For a prespecified coverage target $c^*$, positive scored support yields target support if $L_{A,T}\geq c^*$, undercoverage support if $U_{A,T}<c^*$, and an inconclusive result otherwise. Without scored support the result is not estimable. A wide valid bound is not proof of adequate coverage, and an empirical deficit alone is not a supported sequential undercoverage decision.

Calibration resolution, issued forecasts, scored targets, sensor-outage runs, and time blocks are kept separate. Increasing the number of calibration scores cannot change the audit of already fixed intervals and a fixed evaluation record. Likewise, temporal resampling of native records supplies dependence sensitivity rather than an automatic justification for population inference.

The same accounting supports controlled comparisons. A fixed-model input-mask intervention can vary persistent versus dispersed missingness and predictable timing while preserving original scoring truth. A separate comparison can change the calibration mask while holding evaluation forecasts fixed. Comparing forecasters with a common conformity score, and reporting interval widths, helps distinguish changes in prediction, calibration, and available information. These designs concern supplied input interventions; they do not identify the physical cause of a native outage.

\section{Selected Empirical Evidence}
The underlying evaluation uses public traffic and air-quality records: METR-LA and PEMS-BAY \cite{li2018dcrnn}, AQI-36 and AQI-437 \cite{yi2016stmvl}, and the San Diego subset of LargeST \cite{liu2023largest}. The released availability codes define outages, rather than verified hardware-failure events. For METR-LA and PEMS-BAY, zero-coded readings follow the benchmark's unavailable-value convention; finite zero flow is retained in LargeST. AQI-36 and AQI-437 share stations and overlapping records and are not independent cohorts.

The selected results in Table~\ref{tab:selected} concern the native GRU pipeline with 12 input steps and 12 forecast horizons. These three records use chronological training, tuning, final-calibration, and evaluation partitions of 70\%, 5\%, 5\%, and 20\%, respectively, with windows wholly inside a partition. Normalization uses observed training values. Calibration is separate from model selection. The interval uses the score $|Y-\mu|/\sigma$ and radius $q\sigma$, with the exact augmented split-conformal quantile. Three fixed training seeds share forecast keys; their covered indicators are averaged before pooling originally observed targets. Native temporal dependence means that the nominal level is a calibration target, not an assumed achieved guarantee.

\begin{table}[!t]
\centering
\caption{Selected descriptive coverage at nominal 0.90. Long gaps exceed 12 missing steps at forecast issue time. Values concern originally observed targets and are not confidence bounds.}
\label{tab:selected}
\begin{tabular}{lrr}
\toprule
Record & Marginal coverage & Long-gap coverage\\
\midrule
METR-LA & 0.893 & 0.864\\
AQI-36 & 0.918 & 0.575\\
AQI-437 & 0.919 & 0.793\\
\bottomrule
\end{tabular}
\end{table}

Table~\ref{tab:selected} illustrates why a pooled coverage value is insufficient. The AQI releases have marginal coverage above 0.90 while their long-gap scored subsets fall below it. These selected point estimates are descriptive, not evidence that one network or forecasting method is universally superior. They do not report the complete coverage--width tradeoff.

Scoring availability further limits interpretation. METR-LA has 1,618,824 issued long-gap forecast keys but only 123,099 originally observed scoring targets, approximately 7.6\%. Its measured long-gap coverage therefore describes a small observed subset. Equation~\eqref{eq:identification} makes explicit why this cannot identify coverage for all issued long-gap forecasts.

Across the 30 primary sequential outage-state audits from five records, three outage states, and two nominal levels, 26 results are inconclusive and four lack scored support. None supports the nominal target or undercoverage for the sequential estimand. These outcomes should not be confused with the descriptive deficits in Table~\ref{tab:selected}. They expose the limited precision attainable when the audit allows arbitrary dependence inside whole-network batches. The complete result tables and probability-model validation outputs are not included in this concise presentation.

\section{Limitations and Conclusion}
The audit establishes a connection between issue-time availability, original scoring truth, and a precisely defined evidence state. It does not reconstruct missing truth, prove a population guarantee without its required sampling assumptions, or identify hardware-failure mechanisms from benchmark availability masks. The sequential guarantee concerns \eqref{eq:target}; it is not a guarantee of coverage at every outage state for future populations or for unobserved targets.

Whole-network batching can yield conservative bounds. The inconclusive native results are therefore substantive limits of the evidence, not successful certifications. Independent reference measurements, additional observations under relevant operating states, and event telemetry could address different sources of uncertainty. No universal number of outage events or waiting period follows from this analysis.

Complete forecast accounting and simultaneous count-ratio inference connect measured coverage to an explicit evidence state. This connection makes reliability statements interpretable under incomplete sensing.

\section*{Acknowledgment}
We thank the National Center for High-Performance Computing (NCHC) of National Institutes of Applied Research (NIAR), Taiwan, for computing resources and support. OpenAI ChatGPT and Codex assisted with language editing, organization, literature screening and citation checking, research-code revision, analysis implementation, probability-model experimental design, and automated verification in the underlying study. The reported numerical results were obtained by executing the research programs. The authors are responsible for the scientific content, references, and analyses.

\section*{Data and Code Availability}
The benchmark sources are identified in the references. This concise preprint and its source package contain only the methods and selected aggregate results presented here. They do not include raw records, full experiment tables, forecast-indicator files, model checkpoints, or research code. Those materials remain in the authors' research archive.

\section*{Conflict of Interest}
The authors declare no competing interests.
\bibliographystyle{IEEEtran}
\bibliography{references}
\end{document}

%% file: abstract_text.tex
Sensor outages can conceal prediction-interval coverage deficits behind acceptable marginal performance, while unavailable future readings leave some issued forecasts unscorable. We present a measurement-record audit that links issue-time outage duration, original target availability, and time-uniform statistical evidence. The audit retains complete forecast keys, batches all sensors and overlapping horizons at each measurement tick, and projects simultaneous covered-count and scored-count bounds onto a cumulative conditional expected coverage ratio. This target is distinguished from empirical coverage and outage-conditional population coverage. Selected traffic and air-quality results illustrate the gap between pooled and long-outage coverage; in METR-LA, only 7.6\,\% of issued long-gap forecasts have originally observed targets. Of 30 primary sequential outage-state audits, 26 are inconclusive and four lack scored support. This concise presentation gives the core construction and selected aggregate evidence without the full experimental supplement.